\documentclass{IEEEtran}

\usepackage{filecontents}
\usepackage{amsmath,amssymb,amsfonts}
\usepackage{algorithmic}
\usepackage{graphicx}
\usepackage{textcomp}
\usepackage{xcolor}
\usepackage[most]{tcolorbox}
\usepackage{framed}
\definecolor{shadecolor}{RGB}{180,180,180}

\begin{document}

\title{FootGPT : A Large Language Model Development Experiment on a Minimal Setting\\}

\author{\IEEEauthorblockN{Eren Unlu} \\
\IEEEauthorblockA{\textit{Datategy SAS} \\
Paris, France \\
eren.unlu@datategy.fr}
}

\maketitle

\begin{abstract}

With recent empirical observations, it has been argued that the most significant aspect of developing accurate language models may be the proper dataset content and training strategy compared to the number of neural parameters, training duration or dataset size.  Following this argument, we opted to fine-tune a one billion parameter size pretrained general purpose causal language model with a dataset curated on team statistics of the Italian football league 2017/2018 season’s first ten gameweeks, using low rank adaptation. The limited training dataset was compiled based on a framework where a powerful commercial large language model provides distilled paragraphs and question answer pairs as intended. The training duration was kept relatively short to provide a basis for our minimal setting exploration. We share our key observations on the process related to developing a specific purpose language model which is intended to interpret soccer data with constrained resources in this article.

\end{abstract}


\section{Introduction}

Unequivocally, Large Language Models (LLMs) and generative AI in the broader sense, are one of the most remarkable advancements we have seen in technology in recent years, promising an unimaginable array of potential applications which would transform our future \cite{huang2022towards}. OpenAI’s publicly available ChatGPT’s performance of emulating human conversation and early signs of emerging capabilities sparked the utmost interest of masses at the end of 2022 \cite{schulman2022chatgpt}. Along with their advanced mimicry of human linguistic style, at the heart of these surprising abilities lies the Chain-of-Thought (CoT) mechanism, allowing these models to perform human-like reasoning. In the early phases of examination of emerging capabilities of LLMs, this complex behavior was reserved for relatively very large models (>100B) \cite{wei2022emergent}. However, studies such as \cite{fu2023specializing} have shown that by constraining training (fine tuning) on a specific domain with a properly prepared dataset which conforms well to the intent by including explicit or implicit CoT steps, much smaller models can achieve it as well \cite{fu2023specializing}.  

With the current paradigm it seems that the only viable option to retain advanced features of language models is to trade generic ability for the model size \cite{fu2023specializing}. In order to keep the high quality general statistical linguistic flow, it is preferable to fine-tune a model which is pretrained with very large and diverse corpora for task specialization. New techniques like Low Rank Adaptation (LORA) have opened up the possibility to fine tune LLMs very efficiently whilst not being trapped in the problem of catastrophic forgetting \cite{hu2021lora}. These features made LORA the go-to method for domain specific language model fine-tuning. The temporal and hardware burden of fine-tuning large models were scaled down in exponential scale, where at the same time, thanks to the specific structure of external adaptation, the target specific capabilities can be based on the inherent reasoning and human-like language generation of the main model. In addition to LORA, effective use of quantization further improved computational footprint \cite{dettmers2023qlora}. 

Very large LLMs, some commercially serviced such as ChatGPT, constitute themselves as powerful tools for the fine-tuning process. Most prominent example of this would be the Alpaca model developed by Stanford researchers which made the headlines in language processing recently \cite{taori2023alpaca}. The model is fine-tuned from a 7B Llama (Meta) with only 52,000 instruction-response pairs which are synthetically generated using text-davinci-003 model (Open AI). This particular case shows, from a perspective, how the quality and content of a dataset is much more important than the size and how able but prohibitive LLMs can be utilized in the process of dataset curation. These types of attempts, especially to align purely causal language models (pre-trained to predict tokens in raw corpus) with human instructions are sometimes referred as “distilling” from the teacher model (which synthesizes the dataset and/or evaluates the quality of the student model) \cite{kim2023aligning}, though it may be seen kind of different than the traditional deep learning distillation. 

We opted to investigate the extent of capabilities of a relatively small one billion parameter general purpose causal language model fine-tuned for football data. Since the “moneyball” effect, sports data analysis has revolutionized the sector \cite{lewis2004moneyball}. Though soccer (association football) was relatively late to adopt a data centric approach, perhaps due to its nature of gameplay, since recent years data and machine learning have become indispensable for player scouting, tactical improvement and performance evaluation \cite{weimar2017moneyball}\cite{herberger2021impact}. Hence, we have thought model fine-tuning on the automatic interpretation of this type of data can be an interesting case study for the minimal experiment setting. One particular reason for choosing such a domain is the reliance on structured data, tables of attributes. For instance, \cite{pi2022reasoning} and \cite{wang2023learning} demonstrate the importance of training language models with purely mathematical and structural data that exhibit emergence of the causal reasoning more efficiently. Given the nature of our task, one can refer to the generic definition of the problem in \cite{wang2022survey} in detail. 

The structured data of our experiment is from the publicly available Wyscout \cite{shen2022analyzing} (a football data provider company) dataset for top 5 European leagues of the season 2017/2018, where we have constrained our case to the first ten gameweeks of the Italian league. The dataset includes match statistics of teams and players such as number of specific events, scores etc. and general information about them. For the sake of contracting the domain, our case study includes only teams. 

Our central idea is to curate a linguistic dataset for fine-tuning from these football statistics using a commercially available powerful LLM as a distiller. The details of the intricate distillation and curation framework are given in next sections. For the context of minimal experimental setting, the size of the curated dataset for fine-tuning is kept relatively small, as our main intention is to demonstrate that the quality of the data is the key for the performance.

At the end of the experiment, we aimed for a minimalistic language model aligned with human intentions that can interpret gameplay through the data. Note that, the accuracy and demonstration of capabilities are not the main prospect of this study, where limits of such a domain adaptation are examined. Given the exceedingly prohibitive cost of training LLMs, it is of paramount interest to examine emergence of complex abilities in small language models through domain adaptation based on the aforementioned trade-off. Therefore, we believe our work in this paper can spark new ideas on developing language models with relatively small numbers of parameters for specific tasks accompanied with structured data of similar format with very moderate computational and temporal budget.    

\section{Football Statistics Dataset}

In our exploration, we utilized data sourced from Wyscout \cite{shen2022analyzing}, one of the leading football data providers. For community engagement and research they have published top 5 European leagues’ 2017/2018 stats. Their comprehensive datasets, particularly for the Italian league's 2017/2018 season, provided a fertile ground for our study. For the sake of simplicity with minimal settings, we have only covered the first ten gameweeks of top tier Italian football league, Serie A  as the focal point for our investigation. The dataset in its rawest form is composed of several raw JSON files encapsulating a plethora of details ranging from match outcomes to player-specific in-game events. 

A breakdown of these files and their content is as follows: A json for match scores and details; a comprehensive overview of each game's outcome during the gameweeks under study, final score, the teams competing, line-ups and substitutions with respective minute of the game, impactful events such as cards, goals, the date of the match. A json for in-match events per player; dataset dives deep into the actions of individual players during each game. It catalogues events such as shots taken, tackles made, duels charged and other events. Two json files for team and player information; which gives the details about a team’s competition etc. and a player’s age, team, position, nationality etc.

Before curating our linguistic dataset to fine-tune the model, we have filtered and preprocessed these raw files for our intended purpose. Fig. 1 illustrates the 5 tabular data frames we have generated from the raw files for facilitating the curation process. In order to inform users about the content of the data we fine-tune the language model, we present these intermediate tables : 

Teams info table : a table associates team id to team name and competition.

Players info table : a table giving a player’s id, name, current team, position, age etc.

Aggregated weekly team scores and points table : Each row corresponds to a match and team with match id and team id. Includes opposing team id, gameweek, number of goals scored and points gathered by team id and side (home or away).

Aggregated weekly player scores, minutes, cards : Each row corresponds to a match and player with match id and player id. Includes opposing team id, gameweek, number of goals scored by player, number of yellow/red cards taken by player, whether player has started in first XI, number of minutes player has taken.
Events table : Each row corresponds to an event committed by a player at a match with event id, match id and player id. Includes the event's name, source and destination region of the event on the football pitch.

\begin{figure}[h]
\label{fig:fig_1}
\includegraphics[width=0.99\linewidth]{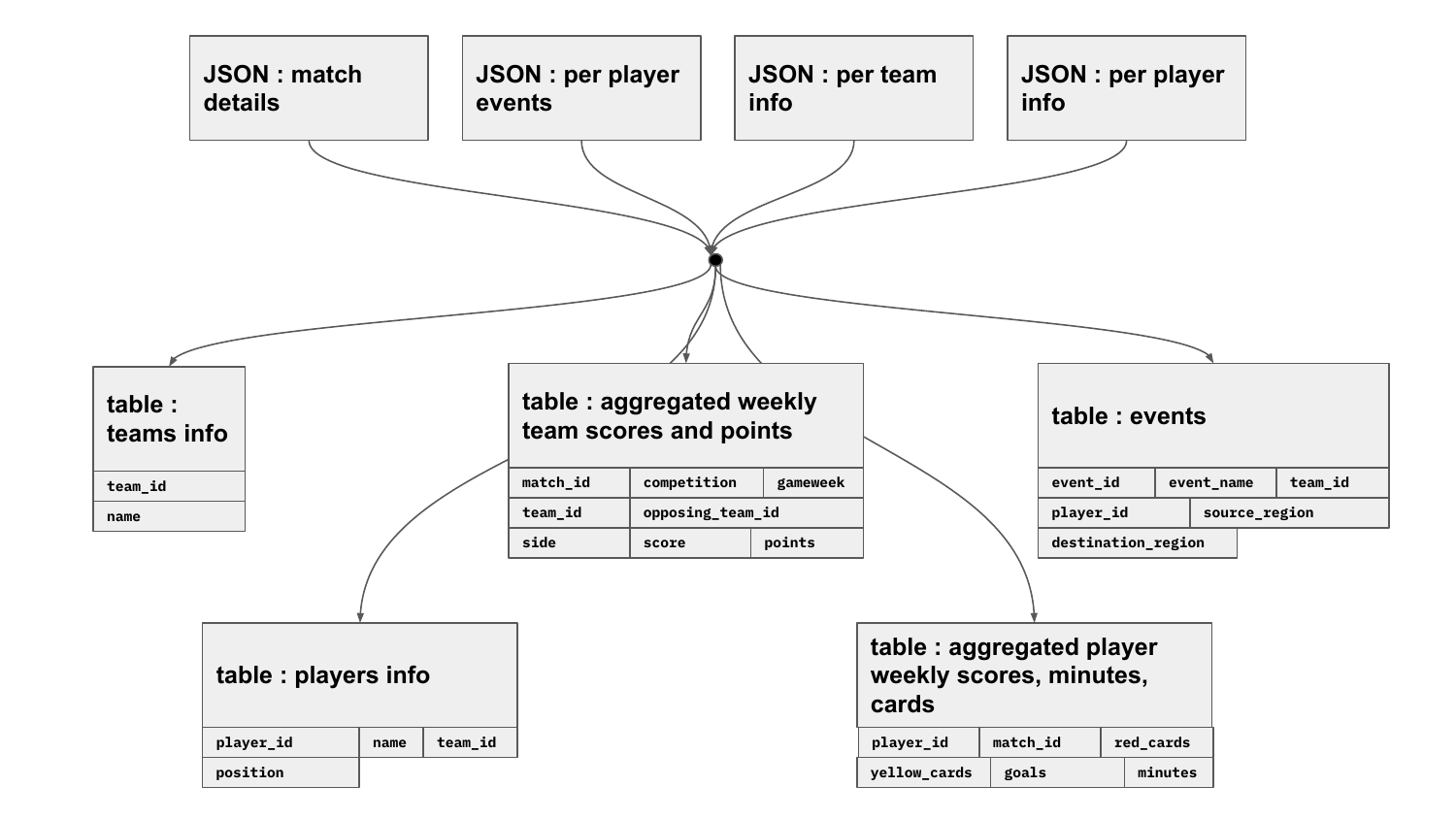}
\caption{5 intermediate tables we generated from raw Wyscout data before curating the linguistic dataset.}
\end{figure}

Without loss of generality, we have determined 11 important types of events to track in this study and excluded the rest :  'Air duel', 'Ground loose ball duel', 'Ground defending duel', 'Ground attacking duel', 'Touch', 'Clearance', 'Shot', Simple pass',  'High pass',  'Head pass' and 'Acceleration'. 

Wyscout dataset also provides the tagged locations of these events on the football field, in X, Y format where it corresponds to location in percentage from top left corner of the field for horizontal and vertical axis, respectively. Certain types of events are defined intrinsically both with source and destination locations : Such as a pass which has a starting location and a delivered destination. 

In order to make the location data comply with the linguistic model, we have divided the field into a rectangular grid of 24 regions as in Fig. 2. To facilitate the fine-tuning process with limited context, we have associated each event with a single location. As certain types of events are defined both with source and destination, we have chosen only one of them. Among them, for instance, ‘Simple pass’ is defined with destination region, ‘Ground loose ball duel’ is defined with source destination etc.

\begin{figure}[h]
\label{fig:fig_2}
\includegraphics[width=0.99\linewidth]{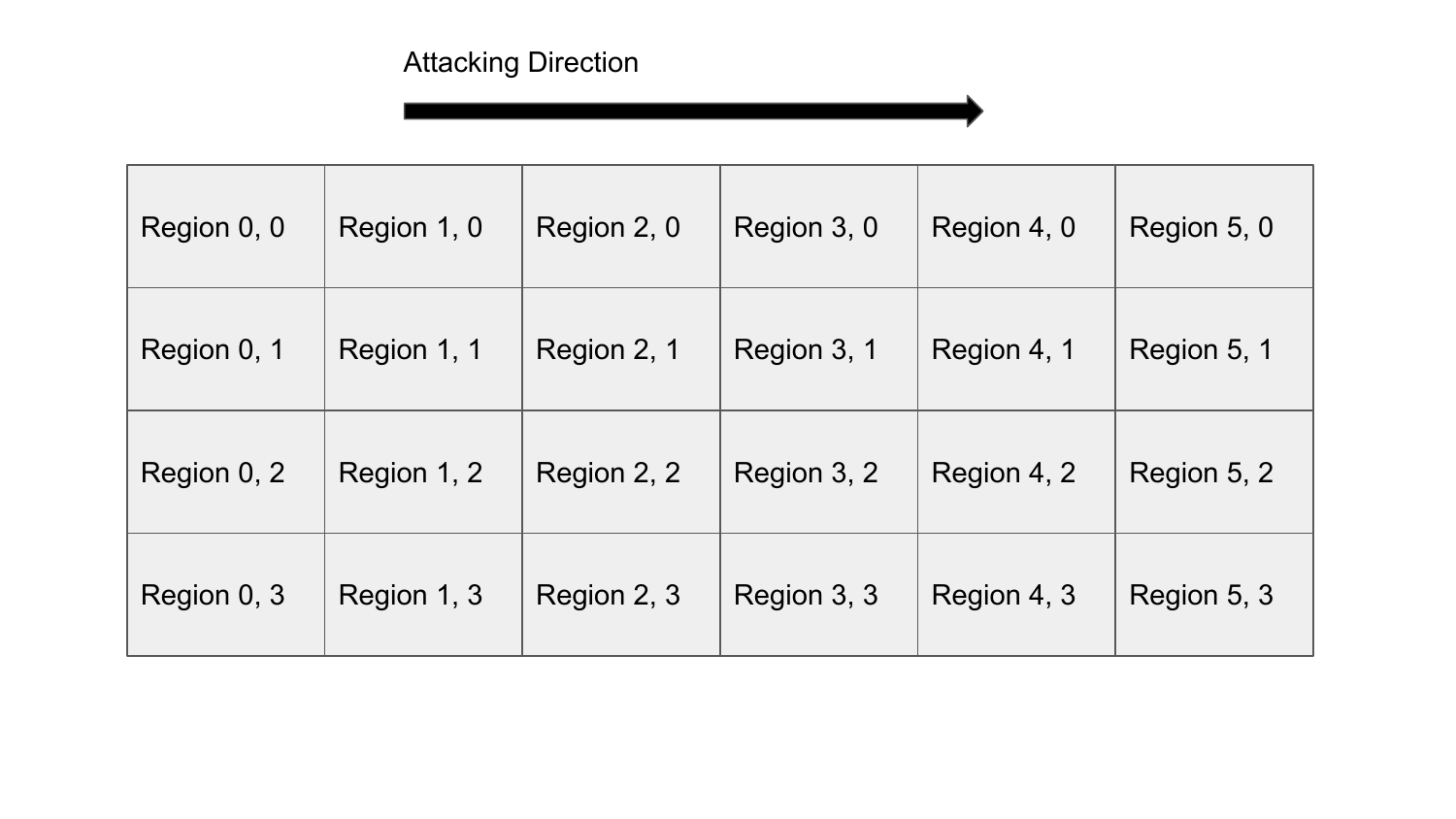}
\caption{Football field is divided into 24 rectangular regions to associate with events.}
\end{figure}

\section{Dataset Curation and Distillation Process}

After filtering, preprocessing and generation of augmented intermediate tables, we start curation of a verbal dataset for the fine-tuning experiment. For most of the cases, we have designed the lines (e.g. prompts from a perspective) as in the question-answer pairs. These questions start always with the prefix “Question : footgpt, …”, where we wanted to instruct the language model with the notion of the oracle like system and the context in this study, that it shall align the pretrained model well more with the data given for fine-tuning. As mentioned previously, we have leveraged a powerful commercially available LLM to phrase templates and synthesize diverse verbal expressions. 

Firstly, we have generated numerous verbal question-answer pairs related to weekly results of matches, current rankings in the league etc. From several handwritten templates both for questions and answers, the powerful teacher LLM is used to rephrase various additional versions. At the end, we could have generated question-answer pairs as follows based on the data : 

\begin{snugshade*}
\noindent\texttt{Question : 'footgpt, where is team Udinese Calcio placed of Italy in gameweek 4?', +Answer : 'In gameweek 4 of league Italy, Udinese Calcio faced off against AC Milan on the away side. The result was Udinese Calcio ended up losing to AC Milan.. Till this gameweek, Udinese Calcio accumulated 3 points, placing them 15 out of 20 teams. In the match, Udinese Calcio netted 1 times against AC Milan.'}
\end{snugshade*}

\begin{snugshade*}
\noindent\texttt{Question : footgpt, how's team AS Roma doing in the Italy during gameweek 3? Answer : Up to this game, AS Roma sits with 4 points and is at position 9 out of 20 teams.}
\end{snugshade*}

\begin{snugshade*}
\noindent\texttt{Question : footgpt, what was Torino FC's goal count on gameweek 10 ? Answer : Torino FC netted 0 times during gameweek 10 when playing ACF Fiorentina.}
\end{snugshade*}

\begin{snugshade*}
\noindent\texttt{Question : footgpt, what was the match result of Genoa CFC played on gameweek 10? Answer : Genoa CFC couldn't hold their ground against SSC Napoli. on gameweek 10 in Italy league.}
\end{snugshade*}

\begin{snugshade*}
\noindent\texttt{Question : footgpt, what's the current standing of team Torino FC in Italy as of gameweek 10? Answer : Torino FC Having collected 13 points by this gameweek, holds the 12 rank among 20 teams}
\end{snugshade*}

\begin{snugshade*}
\noindent\texttt{Question : footgpt, I'd like to know the current position of team SSC Napoli in Italy for gameweek 10. Answer : During gameweek 10 in the Italy, SSC Napoli went head-to-head with Genoa CFC at away. The outcome was SSC Napoli prevailed against Genoa CFC.. Up to this gameweek, SSC Napoli has amassed 28 points, standing 1 among 20 teams. SSC Napoli managed to score 3 goals against Genoa CFC.}
\end{snugshade*}

In order to teach the model the notion of progression of a team by weeks, numerous samples of templates with different parameters (team, start gameweek and finish gameweek) are generated automatically based on the data : 

\begin{snugshade*}
\noindent\texttt{Question : footgpt, from the duration of gameweek 1 to 3, did team SSC Napoli elevate their game? Answer : Team SSC Napoli held the 4th position on gameweek 1. Moving to gameweek 3, it now stands at rank 2. Clearly, an improvement.}
\end{snugshade*}

\begin{snugshade*}
\noindent\texttt{Question : footgpt, during the interval from gameweek 3 to 7, has team Atalanta Bergamasca Calcio stepped up? Answer : Team Atalanta Bergamasca Calcio was positioned at rank 13 during gameweek 3. By the time we reached gameweek 7, it climbed to rank 11. An evident progression.}
\end{snugshade*}

\begin{snugshade*}
\noindent\texttt{Question : footgpt, looking at gameweeks 4 to 6, can you tell if team SS Lazio moved forward? Answer : team SS Lazio's position was 4 For gameweek 4. Jumping to gameweek 6, its rank changed to 5. This is a step backwards.}
\end{snugshade*}

To incorporate the notion of multiple team comparison these types of question-answer pairs are generated from the parameterized templates : 

\begin{snugshade*}
\noindent\texttt{Question : footgpt, based on the matches of the Italian league until gameweek 5, could you evaluate the rankings of SS Lazio, Torino FC, AS Roma, FC Crotone Answer : on gameweek 5 in italy : Torino FC is better than SS Lazio is better than AS Roma is better than FC Crotone}
\end{snugshade*}

\begin{snugshade*}
\noindent\texttt{Question : footgpt, in the Italian league through gameweek 4, how do the rankings of Juventus FC, Bologna FC 1909, US Sassuolo Calcio, Genoa CFC compare? Answer : on gameweek 4 in italy : Juventus FC > Bologna FC 1909 > US Sassuolo Calcio > Genoa CFC}
\end{snugshade*}

\begin{snugshade*}
\noindent\texttt{Question : footgpt, for the match on gameweek 1 between Cagliari Calcio and Juventus FC in Italy league. Which team was better? Answer : Juventus FC was better for this gameweek.}
\end{snugshade*}

\begin{snugshade*}
\noindent\texttt{Question : footgpt, on gameweek 2 in league Italy, can you compare the performances team AC Milan and team US Sassuolo Calcio ? Answer : on gameweek 2 in league Italy, team AC Milan ranks 3 in points and team US Sassuolo Calcio ranks 14. So, AC Milan is better compared to US Sassuolo Calcio. AC Milan scored 5 goals, more (better) than US Sassuolo Calcio which scored 0}
\end{snugshade*}

Note that certain curated prompts may not respect the English orthography and/or logic fully, as we traded simplicity and maximum generic template structures for the accuracy and also wanted to design the content such that the dataset would be better compliant with our simple language model.

We also wanted to introduce arithmetic and statistical cognition up to a degree which is essential for interpreting football data. The fractional numbers are rounded down to a decimal digit of one taking into account the model’s capacity and dataset size. In addition, to further reinforce the concept of temporal progression, cumulative statistics are used in these prompt templates as well.

\begin{snugshade*}
\noindent\texttt{Question : footgpt, till gameweek 9 in league Italy, on average team SS Lazio how many goals scored ? Answer : till gameweek 9 in league Italy, team SS Lazio scored 2.7 goals on average : 24 / 9 = 2.7}
\end{snugshade*}

\begin{snugshade*}
\noindent\texttt{Question : footgpt, till gameweek 5 in league Italy, on average team SSC Napoli how many points collected ? Answer : till gameweek 5 in league Italy, team SSC Napoli has collected 3.0 points on average : 15 / 5 = 3.0}
\end{snugshade*}

Next, the other significant aspect of football statistics is incorporated in dataset curation, the number of events committed by teams in matches. 

The location of the event is crucial for interpretation of gameplay of a team. Therefore, along with number of events per match, the regional decomposition is given in percentages for the top 5 regions frequented for that particular event :  

\begin{snugshade*}
\noindent\texttt{Question : footgpt, considering gameweek 5 in Italy, could you provide the number of High pass team FC Crotone had and their locations on the field? Answer : team FC Crotone on gameweek 5 did 40 High pass(s) / regional decomposition : 25\% in [region 5 , 4]; 17\% in [region 3 , 2]; 17\% in [region 4 , 3]; 15\% in [region 2 , 1]; 10\% in [region 2 , 2]}
\end{snugshade*}

\begin{snugshade*}
\noindent\texttt{Question : footgpt, on gameweek 8 of Italy, can you quantify the Ground loose ball duel involving team FC Crotone and give a pitch-wise breakdown? Answer : team FC Crotone on gameweek 8 did 45 Ground loose ball duel(s) / regional decomposition : 24\% in [region 0 , 0]; 24\% in [region 1 , 1]; 17\% in [region 2 , 1]; 13\% in [region 4 , 3]; 6\% in [region 3 , 2]}
\end{snugshade*}

\begin{snugshade*}
\noindent\texttt{Question : footgpt, on gameweek 3 in league Italy, how many times encounter Simple pass did team Benevento Calcio  and how are these events spread regionally on the field? Answer : team Benevento Calcio on gameweek 3 did 194 Simple pass(s) / regional decomposition : 22\% in [region 4 , 3]; 16\% in [region 3 , 2]; 14\% in [region 2 , 1]; 13\% in [region 1 , 1]; 9\% in [region 2 , 2]}
\end{snugshade*}

Per match averages for events in first ten gameweeks were included as well along with ranking in the league :

\begin{snugshade*}
\noindent\texttt{Question : footgpt, what is the performance ranking of team US Sassuolo Calcio in terms of Shot made per match at the end in league Italy on season 2017/2018 ? Answer : Team US Sassuolo Calcio ranks 15 out of 20 teams in terms of Shot made per match at the end in league Italy on season 2017/2018  (average count : 2.76).}
\end{snugshade*}

\begin{snugshade*}
\noindent\texttt{Question : footgpt, what is the performance ranking of team FC Crotone in terms of Ground defending duel made per match at the end in league Italy on season 2017/2018 ? Answer : Team FC Crotone ranks 1 out of 20 teams in terms of Ground defending duel made per match at the end in league Italy on season 2017/2018  (average count : 22.16).}
\end{snugshade*}

\begin{snugshade*}
\noindent\texttt{Question : footgpt, from gameweek 1 to 10, could you tell the instances of Italy per match for team Shot and their geographical distribution on the pitch? Answer : team US Sassuolo Calcio from gameweek 1 to 10  did 10 Shot(s) per match / regional decomposition : 51\% in [region 4 , 3]; 43\% in [region 5 , 4]; 1\% in [region 3 , 2]; 1\% in [region 5 , 3]; 0\% in [region 3 , 3]}
\end{snugshade*}

More verbal and direct question-answer pairs are also generated to make the model better interpret the football play by leveraging teacher LLM. For this purpose, we have, for instance, gave teacher LLM events of teams in verbal of json format and instructed to curate data for us. Note that, we need to instruct the teacher LLM to understand the notion of defined regions as well.  For this purpose, prompts such as these are effectuated with the teacher model: 

\begin{snugshade*}
\noindent\texttt{given this : team UC Sampdoria from gameweek 1 to 10 did 32 High pass(s) per match / regional decomposition : 34\% in [region 4 , 3]; 22\% in [region 5 , 4]; 15\% in [region 3 , 2]; 8\% in [region 2 , 2]; 6\% in [region 1 , 1]', 'team UC Sampdoria from gameweek 1 to 10 did 395 Simple pass(s) per match / regional decomposition : 22\% in [region 3 , 2]; 16\% in [region 1 , 1]; 16\% in [region 4 , 3]; 14\% in [region 2 , 1]; 12\% in [region 2 , 2]', 'team UC Sampdoria from gameweek 1 to 10 did 48 Touch(s) per match / regional decomposition : 21\% in [region 0 , 0]; 17\% in [region 1 , 1]; 15\% in [region 3 , 2]; 14\% in [region 4 , 3]; 10\% in [region 2 , 1]', … and this : Region 0, 0: Defensive third, near the left corner flag. Region 1, 0: Defensive center, left wing. Region 2, 0: center, left wing. Region 3, 0: center, left. Region 4, 0: Offensive center, left wing. Region 5, 0: Offensive third, left win. …}
\end{snugshade*}

Certain times we instructed the teacher LLM to generate paragraphs out of given data. These paragraphs are used both directly in their raw form in final dataset for fine-tuning and also to re-instruct the teacher to generate question-answer pairs. 

\begin{snugshade*}
\noindent\texttt{write a detailed paragraph about gameplays of teams by mentioning verbally regional decompositions and events}
\end{snugshade*}

For instance, paragraphs like below are produced by the teacher model : 

\begin{snugshade*}
\noindent\texttt{The gameplays of the teams over the gameweeks 1 to 10 provide intriguing insights into their strategic dynamics and on-field tendencies. UC Sampdoria, for example, showed a significant inclination for high passes, averaging 32 per match. Interestingly, most of these passes were focused in [region 4,3] (offensive midfield, right wing) and [region 5,4] (attacking third, central). Moreover, their commitment to maintaining possession is evident from their 395 simple passes per match, predominantly in [region 3,2] (central midfield, left wing) and [region 1,1] (defensive third, far left). Conversely, AC Chievo Verona adopted a multifaceted approach. They engaged in 37 air duels per match, majorly in [region 2,1] (defensive midfield, left wing) and [region 0,0] (goal area, far left). Furthermore, their 305 simple passes each game were concentrated in [region 3,2] and [region 4,3]. Bologna FC 1909's gameplay, characterized by numerous clearances primarily in [region 0,0], indicates a defensive strategy. However, their 79 ground attacking duels per match, often in [region 4,3] and [region 3,2], suggest a contrasting offensive push in the midfield. Finally, Atalanta Bergamasca Calcio displayed a pronounced offensive flair with 90 ground attacking duels each game, with a heavy emphasis on [region 4,3] and [region 5,4] (attacking third, central), indicating a strategy that leans on exploiting the wings and central zones of the offensive third. In essence, the regional decomposition of events offers a nuanced understanding of each team's gameplay strategy, revealing both their defensive and offensive priorities on the pitch.}
\end{snugshade*}

Numerous diverse question-answer pairs such below are generated automatically :

\begin{snugshade*}
\noindent\texttt{Question : footgpt, In which region did AC Milan mostly take their shots? Answer : AC Milan mostly took their shots in regions [4, 3] and [5, 4].}
\end{snugshade*}

\begin{snugshade*}
\noindent\texttt{Question : footgpt, How can the gameplay of Atalanta Bergamasca Calcio be best described? Answer : Atalanta Bergamasca Calcio exhibits significant attacking intent, especially on the wings.}
\end{snugshade*}

\begin{snugshade*}
\noindent\texttt{Question : footgpt, Which team seems to exploit the wings and central zones of the offensive third? Answer : Atalanta Bergamasca Calcio.}
\end{snugshade*}

Many diverse paragraphs and question-answer pairs are generated thanks to the powerful commercial teacher LLM regarding the correlation of number of events with gameplay style and success, interpretation of regional decomposition of given events, general context of football, etc. by instructing prompts such below : 

\begin{snugshade*}
\noindent\texttt{Let's identify and describe each region based on the division of the football field into 6 horizontal and 4 vertical cells: With this perspective, for a football field, the regions are oriented with the team's attack direction. It gives a clear understanding of where each player might be situated relative to offensive and defensive duties during a game …}
\end{snugshade*}

\begin{snugshade*}
\noindent\texttt{generate 50 question answer pairs from this about the gameplay and its relationship to events and regional decomposition for the team (in python list of tuples format ex : [(q,a), (q,a), ..], use " " and add "footgpt, " at the beginning of each question)}
\end{snugshade*}

\begin{snugshade*}
\noindent\texttt{among these football events a team makes, which of them are positevely correlated with performance and which are negatively correlated with performance : 'Air duel', 'Ground loose ball duel', 'Ground defending duel','Ground attacking duel', 'Touch', 'Clearance', 'Shot', 'Simple pass', 'High pass', 'Head pass', 'Acceleration', 'Goals', 'Red Cards', 'Yellow Cards'}
\end{snugshade*}

\begin{snugshade*}
\noindent\texttt{assume football field is 120 (horizontal) by 80 (vertical) meters ... }
\end{snugshade*}

\begin{snugshade*}
\noindent\texttt{given all these above and the json. give conclusions about the playing style of this team through out gameweeks (precising the team name and the opposing teams at each gameweek)}
\end{snugshade*}

\begin{snugshade*}
\noindent\texttt{I will give you the results for the matches gameweek 1 of italian football league in 2017/2018 as a json}
\end{snugshade*}

\begin{snugshade*}
\noindent\texttt{Given that we have number of these football events i will precise for a team during match, write a very detailed paragraph about correlation, relationship etc. between team's performance. I will list the events next.}
\end{snugshade*}

Such paragraphs and question-answer pairs are produced by the teacher model for our curated dataset :

\begin{snugshade*}
\noindent\texttt{In gameweek 3, UC Sampdoria played against AS Roma and scored 1 goal. At that time, UC Sampdoria's rank was 5. Similarly, in gameweek 5, FC Crotone played against Atalanta Bergamasca Calcio and scored 1 goal. Their rank at that time was 19. In the first five gameweeks of the 2017/2018 season in the Italian league, several intriguing matches and results emerged. By the third week, both UC Sampdoria and AS Roma squared off, each securing a goal, showcasing their offensive prowess. FC Crotone, while positioned at 19th, displayed a surprising performance against Atalanta Bergamasca Calcio with a scoreline of 1-5 in favor of Atalanta. It was an astonishing display of dominance by Atalanta. SSC Napoli, a team known for their attacking playstyle, managed to net 4 goals by the fifth gameweek, solidifying their standing. Juventus FC, a consistent top performer in Serie A, maintained their reputation with a decisive victory in one of their matches. Furthermore, the data suggests a varied performance across the teams with several fluctuations in terms of goals scored and their respective rankings. It was a dynamic start to the league with teams jostling for a favorable position early in the season. The dataset provided captures an extensive snapshot of the Italian football league (Serie A) during the first five gameweeks of the 2017/2018 season. The information details various matches that took place among renowned Italian clubs like Juventus FC, AS Roma, AC Milan, FC Internazionale Milano, and others. During these games, a range of goal-scoring outcomes was observed, with some matches even seeing up to 6 goals. The dataset includes not only the names of the participating teams but also specific details such as gameweek, goals scored, and the current rank of the team at that time. It portrays a vivid picture of the competitive nature of Italian football, reflecting the fluctuating successes and struggles of the clubs in the early stage of the season. Whether it was the thrilling 5-goal victory by Atalanta Bergamasca Calcio over FC Crotone in the 5th gameweek or the gripping draw between UC Sampdoria and AS Roma, these initial games set the tone for an exciting season ahead.}
\end{snugshade*}

\begin{snugshade*}
\noindent\texttt{Passing patterns – whether they're 'Simple passes', 'High passes', or 'Head passes' – often paint a vivid picture of a team's offensive blueprint. Fluid, ground-based passing moves can disorient defenses, while a well-timed high ball can exploit the spaces behind defensive lines. 'Acceleration' doesn’t just spotlight a player's physical capability; it represents moments that can change the game's trajectory. A sudden burst of speed can disrupt structured defenses, creating pockets of space that can be exploited.}
\end{snugshade*}

\begin{snugshade*}
\noindent\texttt{The football field, when divided into these specific segments, offers a precise and granular representation of player positioning, tactics, and gameplay dynamics. Starting from the defensive end, [region 0, 0] represents the defensive third near the left corner flag—a critical zone during defending corner kicks and wide threats. Just to its right, [region 1, 0] symbolizes the left wing's defensive center, often patrolled by left full-backs or wing-backs. At the heart of the defense, [region 2, 0] through [region 3, 0] denote the central defense area, where center-backs mostly position themselves. Moving towards the attacking end on the left, [region 4, 0] captures the offensive center on the left wing, while [region 5, 0] signifies the attacking left wing—zones critical for wingers aiming to deliver crosses.}
\end{snugshade*}

\begin{snugshade*}
\noindent\texttt{This dataset provides insights into the early stages of the Italian football league, Serie A, for the 2017/2018 season, covering the first five gameweeks. The teams involved include prominent names such as Juventus FC, AS Roma, SSC Napoli, and AC Milan, among others. Notably, in the third gameweek, UC Sampdoria and AS Roma played against each other, with both teams scoring a goal each. By the fifth gameweek, Atalanta Bergamasca Calcio displayed their offensive prowess by scoring 5 goals against FC Crotone, while SSC Napoli managed to put 6 past Benevento Calcio. This period also showcased varied performances, as teams jostled for positions in the early league table standings. FC Internazionale Milano, Juventus FC, and SSC Napoli were among the standout teams in terms of performance, with many teams showcasing their ambitions for the season ahead. The dataset provides a comprehensive view of the goals scored, opposing teams, and the current rank of teams, offering a snapshot of the league's dynamics during its initial stages. In the initial five gameweeks of the 2017/2018 Italian Serie A season, certain trends and progressions emerged among the competing teams. SSC Napoli, starting strong, was among the top performers with a surge in goal-scoring capability, netting a staggering 6 goals against Benevento Calcio by the fifth gameweek. Their consistency in both attacking and defensive lines saw them climb towards the league's pinnacle, indicating a potential title challenge.}
\end{snugshade*}

\begin{snugshade*}
\noindent\texttt{Question : footgpt, What could be the defensive strength of Udinese Calcio? Answer : Their focus on ground defending duels in regions [0, 0] and [1, 1].}
\end{snugshade*}

\begin{snugshade*}
\noindent\texttt{Question : footgpt, What is Bologna FC 1909's approach in midfield play? Answer : Combative, with a focus on ground duels in both attacking and defending}
\end{snugshade*}

\begin{snugshade*}
\noindent\texttt{Question : footgpt, How does ACF Fiorentina approach their passing? Answer : A balanced approach with a considerable amount of short pass exchanges}
\end{snugshade*}

In addition to these, we wanted to include paragraphs and question-answer pairs in the fine-tuning dataset regarding the system’s self-awareness and cognition of the context : 

\begin{snugshade*}
\noindent\texttt{Question : footgpt, What is footgpt ? Answer : footgpt is the name of the system which is trained with the data, statistics of the football matches (games) played for first 10 gameweeks of italian league. Users can ask you questions about stats and performances of teams in these leagues on season 2017/2018 and get answers.}
\end{snugshade*}

\begin{snugshade*}
\noindent\texttt{Question : footgpt, How many teams in total are in your dataset? Answer : I have data on all teams that participated in league italy during the first ten gameweeks of 2017/2018 season.}
\end{snugshade*}

\begin{snugshade*}
\noindent\texttt{Question : footgpt, can footgpt give answers related to individual players? Answer : No, footgpt can only give answers related to performances of teams in Italy league (first 10 gameweeks) on season 2017/2018.}
\end{snugshade*}

\begin{snugshade*}
\noindent\texttt{At its core, footgpt doesn't just understand rudimentary statistics. It's acquainted with detailed in-game events like 'Air duels', 'Touch', 'Clearance', 'Simple pass', 'Red Cards', and more. Each of these events is more than just a number; they tell a story about the style, strategy, and efficiency of teams on the field.}
\end{snugshade*}

\begin{snugshade*}
\noindent\texttt{Taking it a step further, footgpt possesses the unique capability of regional decomposition. For each match, it provides a granular breakdown of where specific events occurred on the pitch. Want to know in which regions a team made most of its 'High passes' or where the majority of 'Ground defending duels' took place? Footgpt not only gives you the numbers but also highlights the top 5 regions, giving analysts a clear view of a team's spatial strategy and tendencies.}
\end{snugshade*}

\begin{snugshade*}
\noindent\texttt{In essence, footgpt isn't just another statistical tool. It's a comprehensive soccer analysis oracle, bringing the vast football pitch into the realms of data-driven understanding.}
\end{snugshade*}

\begin{snugshade*}
\noindent\texttt{In the heart of Italy's elite football league's data, a new oracle emerges: footgpt. Designed to sift through the thrilling events of the 2017/2018 season's initial 10 gameweeks, footgpt encapsulates a rich array of match outcomes and in-depth gameplay dynamics.}
\end{snugshade*}

After dividing the final corpus into relatively short chunks (that can fit the single context of our small model) we have curated 11,413 datapoints. 

\section{1B Language Model, LORA and Training Process}

As mentioned previously, the main goal of this study is to examine the limits applicability of domain knowledge transfer for football stats interpretation thanks to proper dataset curation, especially, under very minimal settings : constrained model size, training dataset and duration. For this a relatively small but well pretrained causal language model is chosen. We have used “falcon-rw-1b” one billion parameter model, which was pretrained with refined web dataset by also adhering to the principle of better dataset is the key for accuracy in language modeling \cite{penedo2023refinedweb}\cite{jiang2023empirical}. Note that, this LLM does not include any Reinforcement Learning from Human Feedback (RLHF) end nor it was pretrained to align with human instructions. 

It can be said the Low Rank Adaptation (LORA) method has revolutionized the fine-tuning experience in deep learning, especially among the realm of LLM and paved many new avenues for domain adaptation \cite{hu2021lora}\cite{chai2023int2}. LORA bases itself on the observation in \cite{aghajanyan2020intrinsic} that weight matrices of pretrained language models constitute a low rank dimension which explains partly the surprising linguistic efficiency. The authors in \cite{hu2021lora} define weight matrices of an LLM as $W_{0} \in R^{d \times k}$, where the update (fine-tuning) can be simply formulated as $W_{0} + \Delta W = W_{0} + BA$, where  $B \in R^{d \times r}$ , $A \in R^{r \times k}$, and the rank $r << min(d, k)$. So the additional low rank update $\Delta W$ can be very efficiently optimized by only updating A and B, which contain exponentially less number of parameters. In addition to its high computational efficiency, as W is frozen, the catastrophic forgetting is not experienced, where one can fully exploit the linguistic power of the base LLM.

\begin{figure}[h]
\label{fig:fig_3}
\includegraphics[width=0.99\linewidth]{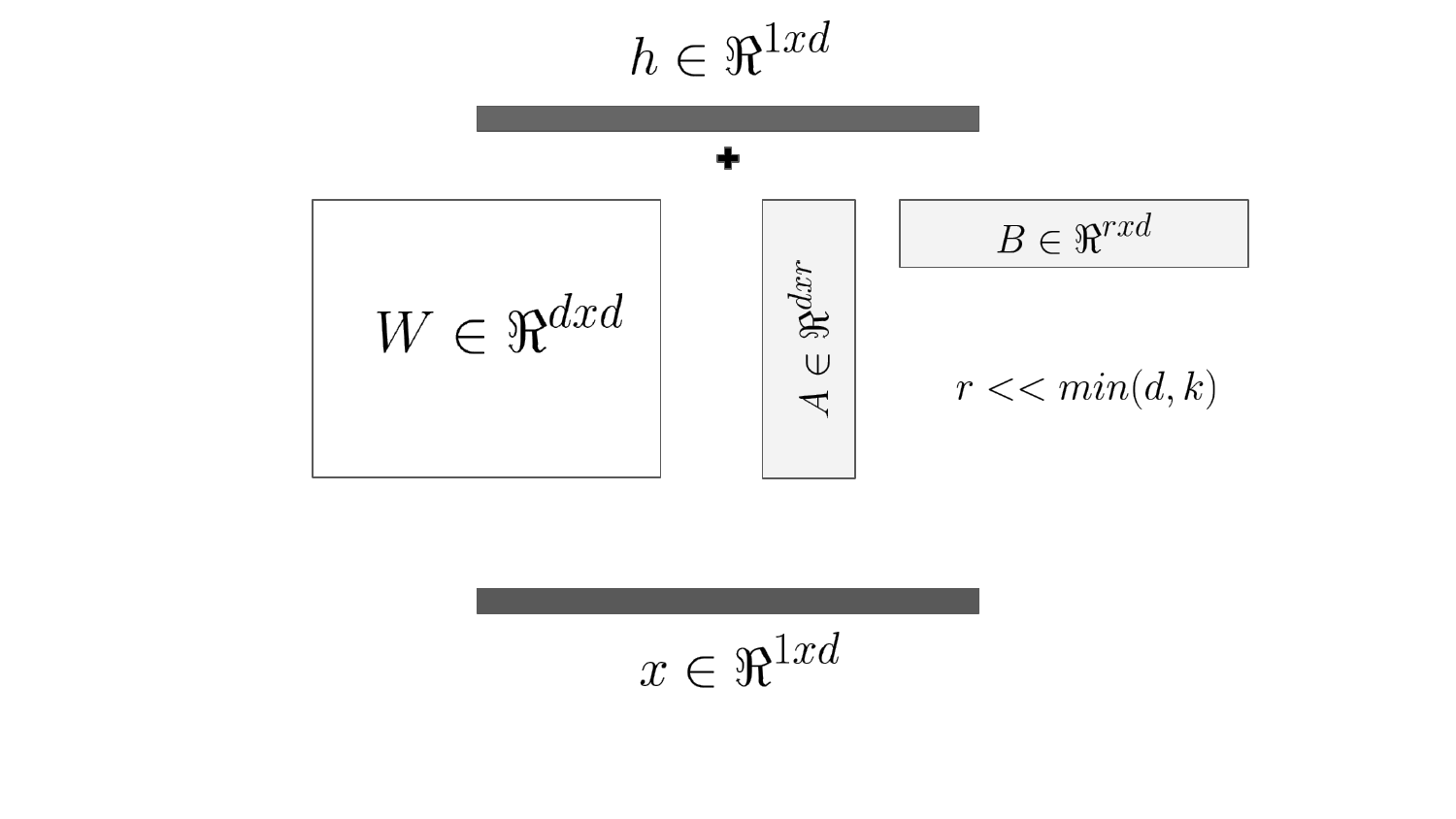}
\caption{Fig. 3 : LORA’s central idea is to inject additive low rank matrices to neural layers to update, whilst freezing the original weight matrix. This allows one to train a substantially lower number of parameters while avoiding catastrophic forgetting \cite{hu2021lora}.}
\end{figure}

Quantization is another important tool to reduce the computational burden of training and deploying LLMs in spite of considerable low accuracy loss \cite{chai2023int2}\cite{liu2023emergent}. Therefore, on top of LORA we have also adopted 8-bit quantization. As mentioned previously, for the sake of minimal setting exploration the base model is fine-tuned only for around 5700 batches which roughly corresponds to 2 epochs. 

\section{Experimental Results}

In this section we present and discuss few of cherry picked responses of the fine-tuned language model (The segments in *\textit{segment}* are prompted to the model and the rest is generated):

\begin{snugshade*}
\noindent\texttt{*Question : footgpt, What differentiates the gameplay of AC Chievo Verona? Answer :* AC Chievo Verona has a balanced gameplay, with a mix of ground attacking and defending duels.}
\end{snugshade*}

\begin{snugshade*}
\noindent\texttt{*Question : footgpt, Does Benevonto Calcio prefers wing play? Answer :* No, Benevonto Calcio prefers central play}
\end{snugshade*}

\begin{snugshade*}
\noindent\texttt{*Question : footgpt, Does Juventus prefer frequent shooting? Answer :* Yes, Juventus prefers frequent shots, especially from the left wing.}
\end{snugshade*}

\begin{snugshade*}
\noindent\texttt{*Question : footgpt, Does Napoli prefer attack from left wing or right wing? Answer :* Team Napoli prefers a balanced approach involving both left and right wings, suggesting a strategic approach involving both wings and wings as well as the center.}
\end{snugshade*}

\begin{snugshade*}
\noindent\texttt{*Question : Which regions of the football field are indicating offense ? Answer :* Regions [0, 0] and [0, 1] of the field are pointing toward an offensive strategy.}
\end{snugshade*}

\begin{snugshade*}
\noindent\texttt{*Question : who is footgpt ? Answer :* footgpt is a tool for analyzing the performance of teams in the Italian league from the italian league season 2017/2018 to the 10th week.
}
\end{snugshade*}

\begin{snugshade*}
\noindent\texttt{*Question : footgpt, which regions of the football field AC Milan prefers do passing more? Answer :* The regions where AC Milan prefers passing more include [4, 3] and [5, 4].}
\end{snugshade*}

\begin{snugshade*}
\noindent\texttt{*Question : footgpt, Which teams in Italian league's 2017/2018 season prefers attacking more on wings? Answer :* SS Lazio and SS Crotone
}
\end{snugshade*}

\begin{snugshade*}
\noindent\texttt{*Question : footgpt, in Italian league 2017/2018, can you tell the name of 3 teams prefering a direct strategy? Answer :* SS Lazio, Genoa CFC, Udinese Calcio}
\end{snugshade*}

\begin{snugshade*}
\noindent\texttt{*Question : footgpt, Does Napoli prefer frequent shooting? Answer :* Yes, Napoli frequently engages in high-quality shots.}
\end{snugshade*}

\begin{snugshade*}
\noindent\texttt{*Question : footgpt, Which wing Napoli prefer to attack more? and why, explain? Answer :* The Napoli's preference to attack on the right side (Region 4, 3).}
\end{snugshade*}

\begin{snugshade*}
\noindent\texttt{*Question : footgpt, Does SSC Napoli shoots less on average per match compared to AC Milan? Give figures. Answer :* Yes, SSC Napoli shoots fewer on average during the match compared to AC Milan.}
\end{snugshade*}

\begin{snugshade*}
\noindent\texttt{*Question : footgpt, given AC Milan did 7 accelerations (mostly in region [4, 3] 57\%) and Calcio did 4 accelerations on the same week (mostly in region [3, 3] 40\%); what can we say about gameplays of these two teams ? Answer :* AC Milan seems more dynamic and more active on the offensive end, whereas Calcio is more likely to rely on quick passes and quick reactions to create spaces in the final third.}
\end{snugshade*}

We have observed many hallucinations and inconsistencies as expected from this minimal setting. However, as it can be seen from the given examples the fine-tuned model shows the glimpses of correct interpretation of football gameplay. By capitalizing on the inherent linguistic capability of 1B base model, it exhibits a highly plausible verbal style on the topic of football. Predictably, the adapter weights are far from memorizing numerical data, however we observe that the curated verbal representation of stats along with proper explaining auxiliary data has induced a strong interpretive capacity. As in the last example, providing stats in-context further augments the capabilities. Note that, it is a well understood phenomenon that especially constrained LLMs can be deemed to mimic more human verbal style rather than possessing imagined cognition \cite{gudibande2023false}.

\section{Conclusion}

In this study, we embarked on the exploration of the limits of domain knowledge transfer for football statistics interpretation using a constrained setting. We curated a comprehensive dataset from football statistics, focusing on the Italian Serie A 2017/2018 season's first ten gameweeks. Through meticulous preprocessing and distillation, we generated a linguistic dataset comprising diverse question-answer pairs and informative paragraphs that encompassed various aspects of the game. The presented framework, especially the proper dataset curation can inspire the community for similar studies. The experimental results unveiled the model's nascent proficiency in interpreting football gameplay, despite the constraints imposed. The cherry-picked responses highlighted the model's ability to grasp elements such as team strategies, preferred play styles, and field regions conducive to different tactics. While certain inconsistencies and hallucinations surfaced as anticipated due to the minimal setting, the responses provided a glimmer of accurate interpretation, indicative of the model's latent potential.


\begin{thebibliography}{9}

\bibitem{huang2022towards}Huang, J. \& Chang, K. Towards reasoning in large language models: A survey. {\em ArXiv Preprint ArXiv:2212.10403}. (2022)

\bibitem{schulman2022chatgpt}Schulman, J., Zoph, B., Kim, C., Hilton, J., Menick, J., Weng, J., Uribe, J., Fedus, L., Metz, L., Pokorny, M. \& Others ChatGPT: Optimizing language models for dialogue. {\em OpenAI Blog}. (2022)

\bibitem{wei2022emergent}Wei, J., Tay, Y., Bommasani, R., Raffel, C., Zoph, B., Borgeaud, S., Yogatama, D., Bosma, M., Zhou, D., Metzler, D. \& Others Emergent abilities of large language models. {\em ArXiv Preprint ArXiv:2206.07682}. (2022)

\bibitem{fu2023specializing}Fu, Y., Peng, H., Ou, L., Sabharwal, A. \& Khot, T. Specializing Smaller Language Models towards Multi-Step Reasoning. {\em ArXiv Preprint ArXiv:2301.12726}. (2023)

\bibitem{hu2021lora}Hu, E., Shen, Y., Wallis, P., Allen-Zhu, Z., Li, Y., Wang, S., Wang, L. \& Chen, W. Lora: Low-rank adaptation of large language models. {\em ArXiv Preprint ArXiv:2106.09685}. (2021)

\bibitem{dettmers2023qlora}Dettmers, T., Pagnoni, A., Holtzman, A. \& Zettlemoyer, L. Qlora: Efficient finetuning of quantized llms. {\em ArXiv Preprint ArXiv:2305.14314}. (2023)

\bibitem{taori2023alpaca}Taori, R., Gulrajani, I., Zhang, T., Dubois, Y., Li, X., Guestrin, C., Liang, P. \& Hashimoto, T. Alpaca: A strong, replicable instruction-following model. {\em Stanford Center For Research On Foundation Models. Https://crfm. Stanford. Edu/2023/03/13/alpaca. Html}. \textbf{3}, 7 (2023)

\bibitem{kim2023aligning}Kim, S., Bae, S., Shin, J., Kang, S., Kwak, D., Yoo, K. \& Seo, M. Aligning Large Language Models through Synthetic Feedback. {\em ArXiv Preprint ArXiv:2305.13735}. (2023)

\bibitem{lewis2004moneyball}Lewis, M. Moneyball: The art of winning an unfair game. (WW Norton \& Company,2004)

\bibitem{weimar2017moneyball}Weimar, D. \& Wicker, P. Moneyball revisited: Effort and team performance in professional soccer. {\em Journal Of Sports Economics}. \textbf{18}, 140-161 (2017)

\bibitem{herberger2021impact}Herberger, T. \& Litke, C. The impact of Big Data and sports analytics on professional football: A systematic literature review. {\em Digitalization, Digital Transformation And Sustainability In The Global Economy: Risks And Opportunities}. pp. 147-171 (2021)

\bibitem{pi2022reasoning}Pi, X., Liu, Q., Chen, B., Ziyadi, M., Lin, Z., Fu, Q., Gao, Y., Lou, J. \& Chen, W. Reasoning like program executors. {\em ArXiv Preprint ArXiv:2201.11473}. (2022)

\bibitem{wang2023learning}Wang, T. \& Lu, W. Learning Multi-Step Reasoning by Solving Arithmetic Tasks. {\em Proceedings Of The 61st Annual Meeting Of The Association For Computational Linguistics (Volume 2: Short Papers)}. pp. 1229-1238 (2023)

\bibitem{wang2022survey}Wang, D., Dou, L. \& Che, W. A Survey on Table-and-Text HybridQA: Concepts, Methods, Challenges and Future Directions. {\em ArXiv Preprint ArXiv:2212.13465}. (2022)

\bibitem{shen2022analyzing}Shen, E., Santo, S. \& Akande, O. Analyzing pace-of-play in soccer using spatio-temporal event data. {\em Journal Of Sports Analytics}. \textbf{8}, 127-139 (2022)

\bibitem{penedo2023refinedweb}Penedo, G., Malartic, Q., Hesslow, D., Cojocaru, R., Cappelli, A., Alobeidli, H., Pannier, B., Almazrouei, E. \& Launay, J. The RefinedWeb dataset for Falcon LLM: outperforming curated corpora with web data, and web data only. {\em ArXiv Preprint ArXiv:2306.01116}. (2023)

\bibitem{jiang2023empirical}Jiang, W., Synovic, N., Hyatt, M., Schorlemmer, T., Sethi, R., Lu, Y., Thiruvathukal, G. \& Davis, J. An empirical study of pre-trained model reuse in the hugging face deep learning model registry. {\em ArXiv Preprint ArXiv:2303.02552}. (2023)

\bibitem{chai2023int2}Chai, Y., Gkountouras, J., Ko, G., Brooks, D. \& Wei, G. INT2. 1: Towards Fine-Tunable Quantized Large Language Models with Error Correction through Low-Rank Adaptation. {\em ArXiv Preprint ArXiv:2306.08162}. (2023)

\bibitem{aghajanyan2020intrinsic}Aghajanyan, A., Zettlemoyer, L. \& Gupta, S. Intrinsic dimensionality explains the effectiveness of language model fine-tuning. {\em ArXiv Preprint ArXiv:2012.13255}. (2020)

\bibitem{liu2023emergent}Liu, P., Liu, Z., Gao, Z., Gao, D., Zhao, W., Li, Y., Ding, B. \& Wen, J. Do Emergent Abilities Exist in Quantized Large Language Models: An Empirical Study. {\em ArXiv Preprint ArXiv:2307.08072}. (2023)

\bibitem{gudibande2023false}Gudibande, A., Wallace, E., Snell, C., Geng, X., Liu, H., Abbeel, P., Levine, S. \& Song, D. The false promise of imitating proprietary llms. {\em ArXiv Preprint ArXiv:2305.15717}. (2023)

\end{thebibliography}
\end{document}